\documentclass[sigconf]{acmart}

\AtBeginDocument{%
  \providecommand\BibTeX{{%
    \normalfont B\kern-0.5em{\scshape i\kern-0.25em b}\kern-0.8em\TeX}}}
\usepackage{enumitem}
\usepackage{booktabs}
\usepackage{multirow}

\copyrightyear{2026}
\acmYear{2026}
\setcopyright{rightsretained}

\acmConference[ICMI '26]{28th ACM International Conference on Multimodal Interaction}{5--9 October 2026}{Napoli, Italy}

\begin{document}

\title{ERR@HRI 3.0 Challenge: Multimodal Detection of Errors and Anticipation in Human-Robot Interactions}

\author{Maria Teresa Parreira}
\email{mb2554@cornell.edu}
\affiliation{%
  \institution{Cornell University}
  \city{Ithaca, NY}
  \country{USA}
}

\author{Micol Spitale}
\affiliation{%
  \institution{Politecnico di Milano}
  \city{Milan}
  \country{Italy}
}

\author{Maia Stiber}
\affiliation{%
  \institution{Microsoft Research}
  \city{Redmond, WA}
  \country{USA}
}

\author{Shiye Cao}
\affiliation{%
  \institution{Johns Hopkins University}
  \city{Baltimore, MD}
  \country{USA}
}

\author{Amama Mahmood}
\affiliation{%
  \institution{Johns Hopkins University}
  \city{Baltimore, MD}
  \country{USA}
}

\author{Chien-Ming Huang}
\affiliation{%
  \institution{Johns Hopkins University}
  \city{Baltimore, MD}
  \country{USA}
}

\author{Hatice Gunes}
\affiliation{%
  \institution{University of Cambridge}
  \city{Cambridge}
  \country{UK}
}

\author{Wendy Ju}
\affiliation{%
  \institution{Cornell University}
  \city{Ithaca, NY}
  \country{USA}
}

\renewcommand{\shortauthors}{Parreira et al.}
\renewcommand{\shorttitle}{ERR@HRI 3.0}

\begin{abstract}
As robots become increasingly integrated into human environments, their ability to detect and respond to errors remains critical for maintaining user trust and interaction quality. While recent advances in machine learning have improved error detection capabilities, most approaches are limited to specific contexts, controlled settings, or pre-extracted features, limiting their generalizability and applicability to real-world conditions. To address this challenge, the third edition of the ERR@HRI Challenge (ERR@HRI 3.0) provided researchers with two complementary datasets that enable end-to-end innovation in methods for both detecting and preventing errors in human-robot interaction.
The challenge offered raw, non-anonymized video data from naturalistic settings: (1) the Bystander Affect Detection (BAD) dataset, containing webcam recordings of 45 participants' spontaneous reactions to robot and human failure scenarios; and (2) the Bad Idea dataset, featuring 29 participants' anticipatory facial responses while predicting action outcomes \textit{before} failures occur. Both datasets were collected via crowdsourcing, capturing the inherent variability of real-world conditions---diverse lighting, camera angles, participant positioning, and environmental contexts. This naturalistic variability, while challenging, provides an authentic testbed for developing robust error detection systems.
Participants developed multimodal machine learning models for bystander reaction detection (Track 1) and anticipatory outcome prediction (Track 2), with an optional cross-dataset generalization track (Track 3). Three teams submitted valid models, all of which surpassed our convolutional neural network baselines. This paper describes the datasets, tasks, baselines, and results of ERR@HRI 3.0, and discusses implications for building generalizable, context-aware, and anticipatory error detection systems for human-robot interaction.
\end{abstract}


\maketitle

\section{Introduction}
\label{sec:intro}

Robot errors -- deviations from expected or intended behavior \cite{honig2018understanding} -- are not merely technical malfunctions but social events that can disrupt interaction flow, diminish user trust, and negatively impact the overall quality of human-robot collaboration \cite{2015salem}. Despite recent advances in machine learning and sensing, autonomous systems still struggle to reliably detect their own mistakes, particularly in the dynamic and unpredictable contexts that characterize real-world deployment.

Error detection in human-robot interaction (HRI) is complicated by the fact that errors manifest in diverse ways, from functional failures such as navigation or manipulation mistakes, to social missteps such as interrupting a user or misreading conversational intent \cite{tian2021taxonomy, 2015mirnig}. Traditional approaches to error detection often rely on task- or domain-specific knowledge that limits generalizability across robots, tasks, and contexts \cite{carlson2005ugvs}, and frequently depend on users to explicitly report a problem, introducing delays that can prevent timely recovery.

A promising alternative leverages the multimodal behavioral signals that humans naturally exhibit in response to unexpected events. Just as people infer errors from bystanders' reactions---confusion, concern, or surprise signaling that something went wrong---robots might detect their own mistakes through observable social cues \cite{bremers2023bystander}. Recent work has demonstrated that user reactions, including facial expressions, speech patterns, and body language, contain information about interaction failures \cite{spitale2023vita,kontogiorgos2021systematic,kontogiorgos2020behavioural,stiber2022effective,stiber2023using,2015mirnig,spitale2023longitudinal, axelsson2024oh,bremers2024usingsocialcuesrecognize}. 

However, most existing approaches have been constrained in two important ways: first, they rely on pre-extracted features rather than raw sensor data, limiting the space of applicable methods; and second, they focus on controlled laboratory settings that may not reflect the variability of real-world deployments. Additionally, these systems tend to focus on \textit{reactive} detection: recognizing that an error has already occurred and elicited an observable response. Anticipating potential errors \textit{before} they fully manifest could instead enable proactive intervention, with meaningful gains for interaction quality and safety. 

The third edition of the ERR@HRI Challenge (ERR@HRI 3.0) addressed these two gaps. Building on two prior editions \cite{spitale2024errhri2024challengemultimodal, cao2025err}, ERR@HRI 3.0 shifted focus along two axes: (1) it introduced \textit{anticipatory} error prediction alongside reactive detection, and (2) it released raw, non-anonymized webcam video rather than pre-extracted features, enabling end-to-end learning directly on visual data and exposing models to realistic sources of variability such as inconsistent lighting, camera angle, and participant positioning. This paper reports on the datasets, tasks, evaluation protocol, baselines, and final results of the ERR@HRI 3.0 Challenge.

\section{Related Work}
\label{sec:related}

\subsection{Robot Error Detection Using Social Signals}
Social signals -- multimodal behavioral cues that convey emotions, intentions, attitudes, and social dynamics -- have long been used to facilitate collaboration in HRI, for example by conveying user preferences, need for help, or engagement breakdowns. Robot errors naturally elicit such signals from users and bystanders \cite{2015mirnig}, and a growing body of work has shown the feasibility of using these elicited reactions to detect robot errors automatically \cite{bremers2023bystander, spitale2023vita, kontogiorgos2020behavioural, kontogiorgos2021systematic, stiber2022effective, stiber2023using}. Nearly all of this work, however, treats error detection as a purely reactive problem: the behavioral signal of interest occurs \textit{during} or \textit{after} the error, rather than in anticipation of it.

\subsection{Benchmarking in HRI}
Benchmarking efforts (shared datasets, tasks, and metrics that let the community compare methods under identical conditions) have played a transformative role in other areas of AI research; ImageNet \cite{imagenet}, for instance, catalyzed rapid, sustained progress in computer vision by giving researchers a common yardstick against which to measure new methods. Sustained, recurring benchmarking efforts of this kind remain comparatively rare in HRI.
 
A few recent efforts have begun to address this gap, though they primarily target the perceptual and reasoning capabilities of large language models (LLMs) and other foundation models in social contexts, rather than error or failure detection specifically. HRIBench \cite{shi2025hribench} benchmarks vision-language models on real-time human perception tasks relevant to HRI, and the Human-Robot Social Interaction (HSRI) dataset \cite{lee2025human} benchmarks foundation models' social reasoning capabilities in human-robot interaction.
 
The ERR@HRI initiative was established as a platform for benchmarking both HRI datasets and multimodal machine learning models for detecting robot errors from behavioral signals. The inaugural edition, ERR@HRI 2024 \cite{spitale2024errhri2024challengemultimodal}, held at ICMI'24, used a dataset of 89 sessions across 23 participants interacting with a robotic well-being coach, with three sub-challenges -- robot mistakes, user awkwardness, and interaction ruptures -- benchmarked using pre-extracted facial, speech, and pose features and an LSTM/BiLSTM/GRU baseline. Building on this, ERR@HRI 2.0 \cite{cao2025err}, held at ACM MM'25, expanded the dataset to 101 sessions across 42 participants interacting with either a social robot \cite{cao2025interruption} or a voice assistant \cite{mahmood2025user} on five collaborative tasks, and separated the detection problem into system-perspective and user-perspective sub-challenges, again benchmarked on pre-extracted multimodal features (581 features spanning facial, audio, and transcribed-speech modalities) with a random-forest baseline.
 
Across both prior editions, submitted models operated exclusively on pre-extracted, low-dimensional feature representations rather than raw sensor streams, and none of the challenge tasks addressed anticipatory or pre-failure behavior. ERR@HRI 3.0 was designed to close both gaps, while continuing the initiative's core mission of benchmarking generalizable, multimodal error detection models in HRI.


\section{The ERR@HRI 3.0 Challenge}
\label{sec:challenge}

ERR@HRI 3.0 provided two complementary datasets, both originally collected via crowdsourcing on Prolific\footnote{\url{https://www.prolific.com/}} in prior studies, and released to challenge participants as raw, non-anonymized webcam video. This design choice served two purposes: it enabled participants to apply modern end-to-end computer vision methods (e.g., CNNs trained directly on pixel data) rather than being restricted to hand-crafted features, and it exposed models to naturalistic sources of variability -- inconsistent webcam resolution, participant positioning, lighting, and background -- that are largely absent from studies conducted in controlled laboratory settings. All the code and instructions can be found on the challenge repository \footnote{\url{https://github.com/IRL-CT/errhri-3-0}\label{repo}}.

\subsection{Datasets}
\label{sec:datasets}

Table~\ref{tab:dataset_comparison} summarizes the two datasets provided in ERR@HRI 3.0.

\begin{table}[t]
    \centering
    \caption{Comparison of the two ERR@HRI 3.0 datasets.}
    \label{tab:dataset_comparison}
    \small
    \resizebox{\columnwidth}{!}{%
    \begin{tabular}{lll}
    \toprule
    \textbf{Characteristic} & \textbf{BAD} & \textbf{Bad Idea} \\
    \midrule
    Participants        & 45              & 29 \\
    Stimulus scenarios  & 46              & 30 \\
    Total video clips   & 1{,}645         & 865 \\
    Total duration      & 25{,}527 s      & 1{,}851 s \\
    Avg. clip length    & $\sim$15.5 s    & $\sim$1.95 s \\
    Frame rate          & 30 fps          & 30 fps \\
    Setting             & Crowdsourced    & Crowdsourced \\
    Data format         & Raw video       & Raw video \\
    Anonymized          & No              & No \\
    Temporal focus      & During failure  & Before failure \\
    Error type          & Observed failure & Anticipated outcome \\
    \bottomrule
    \end{tabular}}
\end{table}

\subsubsection{BAD (Bystander Affect Detection) Dataset}
The BAD dataset \cite{bremers2023bystander} contains webcam recordings of 45 participants' spontaneous facial reactions while watching 46 stimulus videos depicting robot and human failure scenarios (40 failure scenarios, 6 control scenarios with no failure). \textbf{Labels} are binary and derived directly from the stimulus content: \textit{Failure} (1) versus \textit{Control} (0). \textbf{Data format:} raw video files (.mp4) of each participant's face while watching a given stimulus (average clip length $\sim$15.5 s at 30 fps); the corresponding 46 stimulus videos were also released.

\subsubsection{Bad Idea Dataset}
The Bad Idea dataset \cite{parreira2024badidea} captures anticipatory human reactions to scenarios \textit{before} their outcomes are revealed. 29 participants watched 30 action scenario videos (robots or humans performing a task) that were cut off before showing the outcome, and were asked to predict whether the situation would end \textit{well} or \textit{poorly}, while their webcam recorded their facial reactions. \textbf{Labels} reflect the participant's \textit{predicted} outcome -- \textit{Well} (0) or \textit{Poorly} (1) -- rather than the scenario's actual outcome. \textbf{Data format:} pre-extracted video frames (30 fps) of participants watching and reacting to each scenario (average clip length $\sim$1.95 s), together with participants' outcome predictions; the 30 stimulus videos were also released.

\subsection{Training and Test Sets}
\label{sec:splits}

Both datasets were split subject-independently (no participant appears in both the training/validation and test sets) to evaluate generalization to unseen individuals. Table~\ref{tab:splits} summarizes the resulting splits.

\begin{table}[t]
    \centering
    \caption{Dataset splits and label balance.}
    \label{tab:splits}
    \small
    \resizebox{\columnwidth}{!}{%
    \begin{tabular}{lccccc}
    \toprule
    \textbf{Dataset} & \textbf{Split} & \textbf{Part.} & \textbf{Videos} & \textbf{Label 0} & \textbf{Label 1} \\
    \midrule
    \multirow{2}{*}{BAD}      & Trainval & 36 & 1{,}319 & 173 (13.1\%)  & 1{,}146 (86.9\%) \\
                              & Test     & 9  & 326     & 43 (13.2\%)   & 283 (86.8\%) \\
    \midrule
    \multirow{2}{*}{Bad Idea} & Trainval & 23 & 685     & 360 (52.6\%)  & 325 (47.4\%) \\
                              & Test     & 6  & 180     & 103 (57.2\%)  & 77 (42.8\%) \\
    \bottomrule
    \end{tabular}}
\end{table}

The class distributions of the two datasets differ substantially: BAD is heavily imbalanced toward the Failure class (86.9\%), reflecting its stimulus design (40 failure vs.\ 6 control scenarios), whereas the Bad Idea dataset is close to balanced, reflecting the natural variability in participants' subjective outcome predictions.

\subsection{Challenge Tasks}
\label{sec:tasks}

ERR@HRI 3.0 consisted of three tracks, evaluated and ranked independently.

\subsubsection{Track 1: Bystander Reaction Detection (BAD dataset)}
Binary classification of whether a participant is observing a failure versus a control scenario, based solely on their webcam-recorded reaction.

\subsubsection{Track 2: Anticipatory Response Prediction (Bad Idea dataset)}
Binary classification of a participant's predicted outcome (\textit{well} vs.\ \textit{poorly}) from their anticipatory facial behavior, recorded \textit{before} the scenario's outcome was revealed.

\subsubsection{Track 3: Cross-Dataset Generalization (Optional)}
Participants were additionally invited to explore transfer learning and generalization across the two datasets (e.g., training on BAD and testing on Bad Idea), evaluated and awarded separately from Tracks 1 and 2.

\subsection{Evaluation Metrics and Protocol}
\label{sec:evaluation}

Both tracks are binary classification tasks in which each (participant, video) pair is an independent observation. Participants submitted \textit{window-level} predictions, i.e. one row per fixed-length sliding window per clip, with predicted class and probability scores, and declared the frame rate, window size, and slide length used to produce them. Two constraints were enforced on these parameters: the slide could not exceed the window size (both tracks), and, because BAD clips depict an unfolding failure, the window size for Track 1 could not exceed 2 seconds of video regardless of extraction frame rate; Track 2 clips, being inherently short ($\sim$2 s), were exempt from this cap.

\subsubsection{Aggregation to the Video Level}
For Track 1, the video-level prediction was obtained via majority vote across a clip's windows (ties resolved toward \textit{Failure}); for Track 2, the video-level score was the maximum predicted probability for the positive class across a clip's windows, reflecting the fact that a brief, decisive anticipatory signal is more informative than a sustained one for this task.

\subsubsection{Metrics}
Track 1 was ranked primarily by \textbf{macro F1} (video level), with \textbf{balanced accuracy} as a tiebreaker; both were required to be reported given the dataset's strong class imbalance. Track 2 was ranked by \textbf{AUC-ROC} (video level, max aggregation), chosen for being threshold-free and for rewarding models that fire confidently on the relevant anticipatory signal even when it appears only briefly. Secondary metrics reported at both window and video level included F1 for each class, precision, recall, accuracy, and the complementary primary metric of the other track. We additionally reported two temporal metrics computed over positive-class clips only: \textbf{Earliest Detection Time}, the average percentage of a clip elapsed before the first correct window prediction (lower is better), and \textbf{false negative rate (FNR) per video}, the average fraction of a clip's expected windows that miss the positive label (lower is better). For the very short Bad Idea clips, these temporal metrics are best interpreted as measures of prediction consistency across windows rather than of a meaningful detection trajectory.

\subsubsection{Submission Protocol}
Each team could submit predictions on the held-out test set up to three times per track, to limit overfitting to the test set. Participating teams were additionally required to submit a short paper describing their approach, and were strongly encouraged to release their code.

\section{Challenge Baseline}
\label{sec:baseline}

We provided a baseline for each track, built around the \textbf{BadNet} convolutional architecture family (implemented in PyTorch), together with reference data-loading, training, and evaluation code released through the challenge GitHub repository \footref{repo}.

\paragraph{\textbf{Track 1 Baseline: BadNet}}
For Track 1, we trained a BadNet model \cite{bremers2023bystander} directly on 5 fps frames extracted from the raw videos. Table~\ref{tab:baseline_config} (top) lists the selected hyperparameters, chosen via cross-validation on the training/validation split using an inter-participant (subject-independent) protocol.

\paragraph{\textbf{Track 2 Baseline: Fine-Tuned ResNet-34}}
For Track 2, we fine-tuned a ResNet-34 backbone (pretrained on ImageNet) on the 30 fps Bad Idea frames. Table~\ref{tab:baseline_config} (bottom) lists the corresponding hyperparameters.

\begin{table}[t]
    \centering
    \caption{Baseline model configurations.}
    \label{tab:baseline_config}
    \small
    \begin{tabular}{ll}
    \toprule
    \multicolumn{2}{l}{\textbf{Track 1 --- BadNet}} \\
    \midrule
    Activation & sigmoid \\
    Kernel size / base filters & 8 / 64 \\
    Dropout & 0.7 \\
    Learning rate & 0.0001 \\
    Batch size / Epochs & 32 / 350 \\
    Window size / slide & 5 / 2 frames (at 5 fps) \\
    \midrule
    \multicolumn{2}{l}{\textbf{Track 2 --- ResNet-34 (fine-tuned)}} \\
    \midrule
    Dropout & 0.7 \\
    Learning rate & 0.001 \\
    Batch size / Epochs & 64 / 100 \\
    Window size / slide & 10 / 2 frames (at 30 fps) \\
    \bottomrule
    \end{tabular}
\end{table}

Both baselines used a weighted loss to address class imbalance and were trained with inter-participant cross-validation (holding out a distinct set of participants per fold) before a final model was trained on the full training/validation split and evaluated once on the held-out test set.

\paragraph{\textbf{Baseline Results}}
\label{sec:baseline_results}

Table~\ref{tab:baseline_results} reports video-level baseline performance on the held-out test set for both tracks, alongside the temporal metrics.

\begin{table}[t]
    \centering
    \caption{Baseline performance on the held-out test set (video level).}
    \label{tab:baseline_results}
    \small
    \resizebox{\columnwidth}{!}{%
    \begin{tabular}{lccccc}
    \toprule
    \textbf{Track} & \textbf{Primary} & \textbf{F1-macro} & \textbf{Bal. Acc.} & \textbf{Det.\ Time} & \textbf{FNR} \\
    \midrule
    1 (BAD)      & F1 \textbf{0.502}  & 0.502 & 0.504 & 8.8\%  & 0.122 \\
    2 (Bad Idea) & AUC \textbf{0.564} & 0.561 & 0.572 & 35.6\% & 0.384 \\
    \bottomrule
    \end{tabular}}
\end{table}

The Track 1 baseline performed only marginally above chance on macro F1 (0.502), and its per-class F1 scores were highly asymmetric (F1$_{\text{err}}=0.892$, F1$_{\text{cont}}=0.113$), reflecting the dataset's strong skew toward the Failure class -- the model defaults to predicting the majority class far more often than it correctly identifies control scenarios. The Track 2 baseline showed more balanced per-class performance (F1$_{\text{pos}}=0.559$, F1$_{\text{neg}}=0.563$), consistent with the near-balanced label distribution of the Bad Idea dataset, but its AUC-ROC of 0.564 indicates only modest discriminative ability from anticipatory facial behavior alone. The much longer Earliest Detection Time and higher FNR for Track 2 (35.6\% and 0.384, respectively, versus 8.8\% and 0.122 for Track 1) reflect the difficulty of extracting a reliable anticipatory signal from very short clips, rather than a genuine difference in detection latency, and should be interpreted as an index of the baseline's window-to-window consistency rather than a temporal detection trajectory (see Section~\ref{sec:evaluation}).

\section{Participation and Conclusion}
\label{sec:conclusion}

This paper introduced the ERR@HRI 3.0 Challenge, which addressed the problem of multimodal error detection across the temporal spectrum of error management, from anticipatory responses before failures occur to reactive responses during observed failures. A total of 9 teams from 6 countries signed up to participate. Three teams submitted valid models across the challenge's tracks, all surpassing our provided baselines in at least one track.

We aim for the datasets, baseline code, and results from ERR@HRI 3.0 to serve as durable resources for the community; challenge materials will remain available through our GitHub repository for at least three years. Future editions of ERR@HRI will continue to expand the range of interaction contexts, robot embodiments, and temporal phases of error covered by the initiative, with the goal of developing more adaptive, context-aware error detection systems that can operate reliably across the diverse conditions of real-world human-robot interaction.

\begin{acks}
\noindent\textbf{Data access:} The BAD and Bad Idea datasets contain non-anonymized visual data. Challenge participants were required to sign a Data Use Agreement (DUA) prior to receiving access, agreeing to terms including no redistribution rights, use restricted to the purposes of this challenge, and appropriate data security measures.

\noindent\textbf{Open access:} For the purposes of open access, the authors have applied a Creative Commons Attribution (CC BY) license to any Accepted Manuscript version arising.

\noindent\textbf{Funding:} The work of H. Gunes has been supported by CHANSE \& NORFACE through the MICRO project, funded by ESRC/UKRI grant ref.~UKRI572. 
\end{acks}

\bibliographystyle{ACM-Reference-Format}
\bibliography{ref,tese,bibliography}

\end{document}